\documentclass{article}
\usepackage{iclr2027_conference,times}
\usepackage[T1]{fontenc}
\usepackage{amsmath,amssymb,booktabs,array,graphicx}
\usepackage[hidelinks]{hyperref}
\usepackage{xurl}
\usepackage{float}
\usepackage{wrapfig}
\iclrfinalcopy
\title{PWM: Personalized World Models with\\
Online Reinforcement Learning}
\author{Zhexin Lou$^{1*}$~~
Guancheng Lu$^{2*}$~~
Zeyu Zhang$^{1*\dag}$~~
\textbf{Yi Zhang}$^{1}$~~
\textbf{Yang Zhao}$^{3}$~~~~
\textbf{Hao Tang}$^{1\ddag}$\\[0.5em]
$^1$School of Computer Science, Peking University~~
$^2$Northwestern University~~
$^3$La Trobe University\\[0.4em]
\small $^*$Equal contribution.
$^\dag$Project lead.
$^\ddag$Corresponding author: bjdxtanghao@gmail.com.
}
\begin{document}
\raggedbottom
\maketitle
\lhead{\small PWM: Personalized World Models with Online Reinforcement Learning}
\begin{figure}[H]
    \centering
    \includegraphics[width=\textwidth]{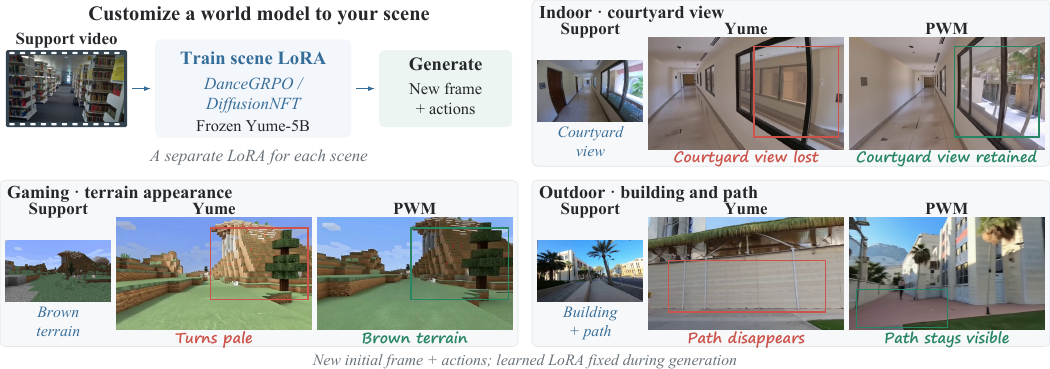}
    \caption{\textbf{Scene customization with PWM.}
        A short support video trains a scene LoRA on a frozen world model.
        Results use DanceGRPO. Indoor and Outdoor show matched generation
        times; Gaming shows different steps at similar viewing distances.}
    \label{fig:teaser}
\end{figure}

\begin{abstract}
Pretrained world models can generate diverse environments,
yet users often want to explore a particular scene specified
by their own video.
This requires learning the scene's visual identity while
retaining the quality of action-conditioned generation.
We introduce Personalized World Models (PWM), a framework
for customizing interactive world models from short scene
videos through online reinforcement learning.
In PWM, the support trajectory and its associated controls
provide reward feedback on continuations sampled from
the current policy.
In the GRPO instantiation, group-relative optimization
updates a compact LoRA adapter using a unified reward
for scene appearance, visual continuity, and motion,
while base-policy anchoring regularizes changes to the
pretrained generation prior of a frozen Yume-5B backbone.
The same adaptation procedure is applied across real
and rendered environments.
We also instantiate PWM with DiffusionNFT as an alternative
reward-guided optimization method for learning the
scene-specific adapter.
We also introduce PWM-Bench, comprising 150 customization
tasks across Indoor, Outdoor, and Gaming, with paired
evaluation on held-out continuations.
The GRPO and DiffusionNFT instantiations of PWM improve
customization over native Yume in 71.3\% and 65.3\%
of the evaluated scenes, respectively, with positive
mean gains across all three domains.
For the GRPO instantiation, matched SFT comparisons
further demonstrate higher mean customization gains
and better mean image-quality scores in every domain,
while retaining frame-level visual quality close
to the pretrained model.
Code: \url{https://github.com/AIGeeksGroup/PWM}.
Website: \url{https://aigeeksgroup.github.io/PWM}.
\end{abstract}
\clearpage

\section{Introduction}
\label{sec:intro}

Interactive world models enable users to explore environments
through action-conditioned video generation.
Recent systems combine pretrained visual generation priors
with user controls and autoregressive rollout, supporting
interactive exploration across diverse
environments~\citep{gamefactory,mao2025yume15}.
Personalizing this experience introduces a concrete
requirement: the generated world should retain the identity
of the environment provided by the user.
For a particular interior, outdoor location, or game scene,
its characteristic appearance and content should remain
recognizable as the generated trajectory progresses.

Personalization methods have demonstrated that pretrained
generators can acquire visual identity from limited
reference observations~\citep{ruiz2023dreambooth}.
Interactive world generation extends this objective to
action-conditioned continuations.
A short support video provides observations of a scene's
materials, lighting, visual style, and characteristic
content.
Successful customization should therefore strengthen
fidelity to the target environment while retaining
the visual quality of the underlying generator.

Reward-based optimization provides a way to express
preferences directly on generated images and
videos~\citep{xue2025dancegrpo}.
For scene customization, the support trajectory can
supply the reference needed to evaluate those outputs:
whether they match the environment's appearance and
whether their visual evolution is consistent with
the observed trajectory.

We introduce \textbf{Personalized World Models (PWM)},
a framework for per-scene customization through online
reinforcement learning.
Given a short support trajectory and its controls,
PWM learns a compact LoRA adapter on a frozen Yume-5B
backbone.
The current policy samples groups of candidate
continuations, and reward feedback derived from the
support trajectory guides updates to the scene-specific
adapter.
We instantiate this framework with GRPO and DiffusionNFT
as alternative reward-guided optimization methods.
Online learning here refers to collecting fresh
rollouts from the current policy during adaptation.

Both instantiations use native generation windows, the same
adapter capacity, and a shared reward for scene appearance,
visual continuity, and motion.
The GRPO instantiation additionally uses a base-policy anchor.
Optimization and checkpoint-averaging settings are fixed
across scenes within each instantiation.

To evaluate this capability, we introduce
\textbf{PWM-Bench}, a suite of 150 per-scene customization
tasks spanning Indoor, Outdoor, and Gaming.
Each task provides an 80-frame support segment and
a temporally later, non-overlapping 145-frame test
continuation.
The benchmark evaluates scene fidelity through
appearance agreement and an entity-based hallucination
penalty, complemented by frame-level image-quality
measurements.
Comparisons use three paired random seeds per scene
under shared initial context, controls, and generation
settings.

Across PWM-Bench, the GRPO instantiation of PWM improves
customization over native Yume in 107 of 150 scenes,
with a mean relative gain of 6.39\%.
The DiffusionNFT instantiation improves 98 of 150 scenes,
with a mean relative gain of 5.22\%.
Both instantiations achieve positive mean gains
in every domain.
Compared with matched SFT, the GRPO instantiation
increases the fraction of scenes improved over Native
from 57.3\% to 71.3\% and achieves higher mean
customization gains across all three domains.
It also obtains better domain-level means than SFT
on MUSIQ, NIQE, and BRISQUE, while remaining close
to native Yume in these frame-level quality measures.

Our contributions are threefold:
\begin{itemize}
    \item
    We introduce PWM, a per-scene customization framework
    that learns a compact adapter from a short support
    trajectory through online reinforcement learning
    on a frozen interactive world model, instantiated
    with GRPO and DiffusionNFT.

    \item
    We introduce PWM-Bench, comprising 150 customization
    tasks across three first-person domains, with
    held-out continuations and a shared protocol
    for evaluating scene fidelity and visual quality.

    \item
    We demonstrate positive average customization gains
    over native Yume across all three domains with
    both optimization methods.
    Matched SFT comparisons show that the GRPO
    instantiation benefits more scenes and achieves
    better mean image-quality scores.
\end{itemize}

\section{Related Work}
\label{sec:related}

\paragraph*{Interactive world models.}
Interactive world generation combines visual synthesis with action
control to produce trajectories that respond to user inputs.
GameFactory~\citep{gamefactory} learns action control from game videos
and uses a domain adapter with multi-phase training to transfer this
capability to open-domain scenes.
Yume~\citep{mao2025yume} combines camera control with autoregressive
video generation, while Yume-1.5~\citep{mao2025yume15} develops context
compression and text-controlled interaction for continuous world
exploration.
These advances establish a foundation for generating and navigating
diverse environments.
PWM studies how to specialize such a pretrained model to a particular
observed scene.
Given a short support trajectory, it learns a compact scene adapter
for subsequent action-conditioned generation in the same environment.

\paragraph*{Personalization of generative models.}
Personalization adapts a generative prior to visual content specified
by limited reference observations.
DreamBooth~\citep{ruiz2023dreambooth} learns a subject from a few images
and synthesizes that subject in new contexts.
Tune-A-Video~\citep{wu2023tuneavideo} extends adaptation from limited
observations to video generation by tuning an image diffusion model
on a single text--video pair.
Parameter-efficient methods such as LoRA~\citep{hu2021lora} support
this adaptation through trainable low-rank updates to frozen
pretrained weights.
PWM applies scene-level personalization to an interactive world
model: the support video provides observations of the target
environment, and the learned adapter is evaluated on a temporally
held-out continuation under action controls.

\paragraph*{Reinforcement learning for visual generation.}
Reward-based fine-tuning adapts visual generators to objectives
defined on their outputs.
DDPO~\citep{black2024training} treats iterative denoising as a
sequential decision process and applies policy gradients to optimize
sample-level rewards.
VADER~\citep{prabhudesai2024vader} optimizes video generation by
propagating gradients from differentiable reward models.
DanceGRPO~\citep{xue2025dancegrpo} applies group-relative policy
optimization to diffusion and rectified-flow models across image
and video generation.
DiffusionNFT~\citep{zheng2026diffusionnft} performs online
reward-guided optimization through the forward noising process,
incorporating positive and negative learning signals into a
flow-matching objective without requiring likelihood estimation.
Building on this line of work, PWM obtains reward feedback from
the target scene's support trajectory and updates a scene-specific
adapter using online rollouts.
We instantiate PWM with GRPO and DiffusionNFT, using the same
scene-level task, adapter capacity, and reward definition.
The reward combines visual agreement with initial-frame consistency
and motion magnitude matching, connecting reward-based optimization
to the scene customization objective.

\paragraph*{Weight averaging.}
Weight averaging combines parameter states into a single model.
Stochastic Weight Averaging~\citep{izmailov2018swa} averages weights
along a training trajectory, while
Model Soups~\citep{wortsman2022modelsoups} averages models fine-tuned
from a shared pretrained initialization.
PWM uses this idea within each scene's adaptation run, averaging
adapter checkpoints to retain a single compact adapter for inference.
Each optimization instantiation uses a checkpoint-averaging rule
that is fixed across scenes and does not use held-out test frames.

\section{Method}
\label{sec:method}

We propose PWM, a per-scene personalization approach for
interactive world models. Given a short preceding trajectory,
PWM trains a scene-specific LoRA on frozen Yume-5B to transfer
scene knowledge to a subsequent trajectory. The core objectives
are visual identity tailored to the target scene, persistent
spatial layout, and action responsiveness, while retaining the
general action and control priors of the pretrained world model.
For this reason, PWM combines scene-specific low-rank
parameterization, online reward-guided optimization with GRPO
or DiffusionNFT based on native generation windows, a unified
reward that balances scene fidelity and motion, and a weight
averaging strategy that does not require selecting a single
checkpoint.

\begin{figure}[t]
    \centering
    \includegraphics[width=\textwidth]{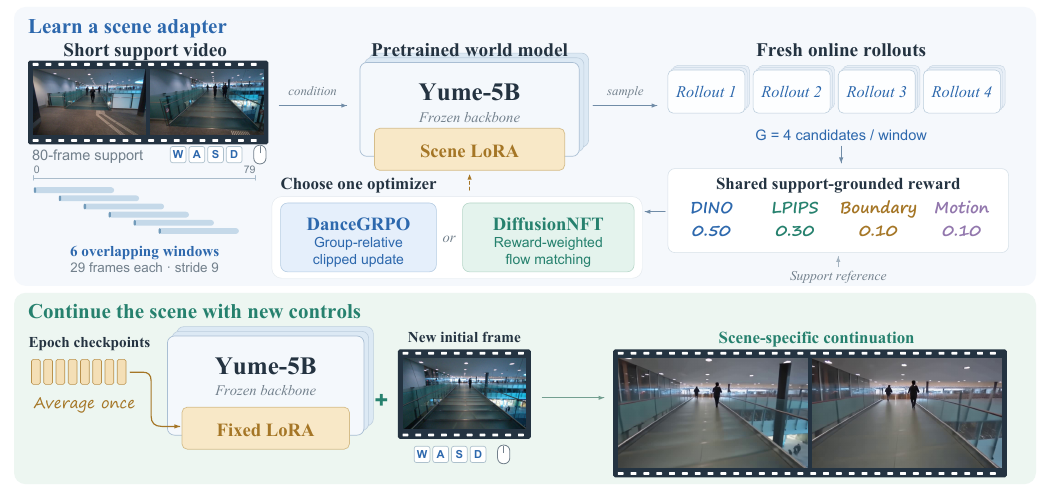}
    \caption{\textbf{PWM with GRPO and DiffusionNFT.}
    Support video and controls condition fresh rollouts from frozen Yume-5B
    with a trainable scene LoRA. Rewards guide GRPO's clipped objective
    with a base-policy anchor, or DiffusionNFT's reward-weighted positive
    and negative flow-matching branches. Only LoRA parameters are updated.
    Eight epoch adapters are averaged for generation.
    Rollout icons are schematic.}
    \label{fig:pwm-method}
\end{figure}

Figure~\ref{fig:pwm-method} summarizes the shared adaptation pipeline
and the two online optimization instantiations.

\subsection{Preliminaries: Yume, GRPO, and DiffusionNFT}
\label{sec:preliminaries}

\paragraph*{Yume interactive world model.}
Yume-1.5~\citep{mao2025yume15} extends a pretrained video generator
into an action-command-controllable interactive world model.
Given a compressed visual history $h_t$ and a sequence $u_t$
containing action commands for generation window $t$, the frozen
Yume-5B model, parameterized by $\theta_0$, generates the next
video segment according to
\begin{equation}
    X_t \sim p_{\theta_0}(X_t \mid h_t, u_t).
\end{equation}
The control sequence combines discrete translation and camera
commands with continuous motion magnitudes. Each generated
segment is incorporated into the visual history to condition
the subsequent window, resulting in a segment-level autoregressive
interaction trajectory. PWM keeps $\theta_0$ frozen and learns
only a scene-specific low-rank adapter.

\paragraph*{Group Relative Policy Optimization.}
Group Relative Policy Optimization (GRPO) estimates advantages
by comparing a group of samples generated under the same
condition, thereby eliminating the need for a learned
critic~\citep{shao2024deepseekmath}. Given a condition $c$,
GRPO samples $G$ trajectories from an old-policy snapshot
and normalizes their outcome rewards within the group:
\begin{equation}
    \tau_i \sim
    \pi_{\theta_{\mathrm{old}}}(\cdot\mid c),
    \qquad
    \widehat{A}_i =
    \frac{r_i-\overline{r}}{s_r},
    \qquad i=1,\ldots,G,
    \label{eq:grpo-advantage}
\end{equation}
where $\overline{r}$ and $s_r$ denote the mean and standard
deviation of $\{r_j\}_{j=1}^{G}$. Under outcome-level supervision,
the same $\widehat{A}_i$ is assigned to every decision step
along trajectory $\tau_i$.

Let $K_i$ be the number of decision steps in $\tau_i$.
GRPO optimizes the empirical clipped surrogate
\begin{equation}
\begin{split}
    \widehat{J}_{\mathrm{clip}}(\theta)
    &=
    \frac{1}{G}\sum_{i=1}^{G}\frac{1}{K_i}
    \sum_{k=1}^{K_i}
    \min\Bigl(
        \rho_{i,k}(\theta)\widehat{A}_i,\,\\
    &\qquad\qquad
        \operatorname{clip}\bigl(
            \rho_{i,k}(\theta),
            1-\varepsilon_{\mathrm{clip}},
            1+\varepsilon_{\mathrm{clip}}
        \bigr)\widehat{A}_i
    \Bigr),\\
    \rho_{i,k}(\theta)
    &=
    \frac{\pi_{\theta}(a_{i,k}\mid s_{i,k})}
         {\pi_{\theta_{\mathrm{old}}}(a_{i,k}\mid s_{i,k})}.
    \label{eq:grpo-objective}
\end{split}
\end{equation}
GRPO therefore replaces a learned critic with a group-reward
baseline rather than removing the baseline altogether.
Its original formulation additionally regularizes the policy
toward a separate reference policy $\pi_{\mathrm{ref}}$.
The GRPO instantiation of PWM instead employs the base-policy
anchor introduced in Section~\ref{sec:method-grpo}.

\paragraph*{Policy interpretation of diffusion and flow sampling.}
Following DDPO and DanceGRPO~\citep{black2024training,xue2025dancegrpo},
iterative diffusion and rectified-flow sampling can be formulated
as a finite-horizon Markov decision process.
Let $c_t=(h_t,u_t)$ denote the visual history and control
condition for video window $t$. After stochasticizing the
reverse-time transition, the state, action, and policy at
denoising step $k$ are defined as
\begin{equation}
\begin{gathered}
    s_{t,k}=(c_t,k,z_{t,k}),\qquad
    a_{t,k}=z_{t,k-1},\\
    \pi_\theta(a_{t,k}\mid s_{t,k})
    =
    p_\theta(z_{t,k-1}\mid z_{t,k},c_t).
    \label{eq:denoising-mdp}
\end{gathered}
\end{equation}
A complete latent trajectory
$\tau_t=(z_{t,K},\ldots,z_{t,0})$ produces one video segment
and receives a terminal reward. Under group-relative
optimization, the resulting trajectory advantage is assigned
to each denoising transition. Here, $k$ indexes the internal
denoising process, whereas $t$ indexes the external sequence
of autoregressive video windows.

We use \emph{online} to mean that each data-collection round
collects fresh trajectories from the latest behavior-policy
snapshot. It does not refer to continual adaptation after
deployment.

\paragraph*{DiffusionNFT.}
DiffusionNFT~\citep{zheng2026diffusionnft} performs online
reward-guided optimization through a forward-process
flow-matching objective.
Generated samples are assigned reward-dependent weights,
which determine their contributions to positive and negative
learning branches.
These branches are constructed from the current model's
velocity prediction and a fixed behavior-policy prediction,
allowing reward-guided updates without estimating
sampling-trajectory likelihoods.
PWM instantiates this objective for scene-specific LoRA
adaptation, as described in Section~\ref{sec:method-nft}.

\subsection*{Problem Formulation: Per-scene Customization}
\label{sec:method-task}

Let a scene $s$ be a temporally ordered controlled trajectory
$\mathcal{D}_s=\{(x_t,u_t)\}_{t=1}^{T}$, where $x_t$ is a video
frame and $u_t$ contains translation, camera, and continuous
motion controls. A support prefix
$\mathcal{D}_s^{\mathrm{sup}}$ is available for adaptation.
The goal is to specialize the model to the persistent identity
of $s$ and transfer that knowledge to a later, unseen portion
of the same scene. No frame from the future evaluation portion
is used for optimization or checkpoint construction.

We start from a frozen Yume-5B world model with parameters
$\theta_0$. For each scene, PWM learns an independent rank-$8$
LoRA adapter $\Delta\theta_s$ (20.2M trainable parameters),
while all backbone parameters remain fixed.
The resulting policy is $\pi_{\theta_0+\Delta\theta_s}$.
This factorization assigns complementary roles to the two
parameter sets: the frozen backbone supplies the general
action-conditioned generation prior, whereas the adapter
stores knowledge specific to the target scene.
The current paper uses the same adapter capacity for both
optimization methods, every scene, and every first-person domain.

\subsection{Per-scene Customization via Online GRPO}
\label{sec:method-grpo}

\paragraph*{Learning in the deployment geometry.}
PWM trains directly with Yume's native $29$-frame generation
window. We cover the support trajectory with six overlapping
windows whose start indices are $[0,9,18,27,36,45]$.
Each window is initialized by its real first frame and is
conditioned on the controls measured for that window.
Training and deployment therefore share the same temporal
geometry; no auxiliary short-window policy is introduced.

\paragraph*{On-policy group-relative optimization.}
PWM instantiates the GRPO update in
Eq.~\ref{eq:grpo-objective} with a group of $G=4$ diffusion
trajectories for each support window.
The trajectories share the initial latent noise to control
nuisance variation while retaining independent stochastic
SDE transitions. Because the unified reward contains components
with different numerical scales, we normalize each component
separately before combining their advantages.
For reward component $m$ and rollout $i$, we compute
\begin{equation}
    A_i = \sum_m \alpha_m
    \frac{R_{i,m}-\mu_m}{\sigma_m+\varepsilon},
    \qquad
    (\mu_m,\sigma_m)
    =
    \operatorname{MeanStd}_{j=1}^{G}
    \!\left[R_{j,m}\right],
    \label{eq:component-advantage}
\end{equation}
where $\alpha_m$ is the weight of component $m$
in Eq.~\eqref{eq:reward}.
The resulting $A_i$ replaces the scalar-reward advantage
$\widehat{A}_i$ in Eq.~\eqref{eq:grpo-objective}.
This component-wise normalization prevents a reward term
from dominating the policy update merely because of its
numerical scale.

We use $\varepsilon=10^{-6}$ and set a component's
contribution to zero when its within-group standard
deviation is below $10^{-3}$
(Appendix~\ref{app:advantages}).

In the PWM implementation, the clipped surrogate uses
$\widetilde{\rho}_{i,k}=\rho_{i,k}^{1/d}$, computed from
Gaussian log densities averaged over the $d$ scalar
latent elements in the generated region
(Appendix~\ref{app:advantages}).

\paragraph*{Noise-robust score-on-mean.}
In few-step diffusion sampling, the final stochastic transition
can inject visible noise into the decoded rollout.
Scoring that sample directly can make the relative advantage
reflect the last noise draw rather than scene content.
PWM therefore separates the trajectory used for policy learning
from the video used for reward measurement.
For the final transition
\begin{equation}
    z_i' = \mu_{\theta}(z_i,c_s) + \sigma\epsilon_i,
    \label{eq:sde-transition}
\end{equation}
the policy likelihood is evaluated at the sampled state $z_i'$,
but the reward is evaluated on the decoded conditional mean
$\widetilde{x}_i=
\operatorname{Dec}(\mu_{\theta}(z_i,c_s))$.
Thus, exploration remains stochastic and on-policy, while
rollout ranking is based on a noise-reduced estimate of
scene content.

Finally, the GRPO instantiation of PWM anchors every updated
transition to the frozen base model.
Let $\mu_{\theta}(s_{i,k})$ and $\mu_{\theta_0}(s_{i,k})$
denote the transition means predicted with the adapter enabled
and disabled, respectively, at the same denoising state.
The GRPO objective used in PWM is
\begin{equation}
\begin{aligned}
    \mathcal{L}_{\mathrm{PWM}}
    &=
    -\widehat{J}_{\mathrm{clip}}(\theta_0+\Delta\theta_s)\\
        &\quad+
    \frac{\lambda}{G}\sum_{i=1}^{G}\frac{1}{K_i}
    \sum_{k=1}^{K_i}
    \frac{1}{d}\left\|
        \mu_{\theta_0+\Delta\theta_s}(s_{i,k})
        -\mu_{\theta_0}(s_{i,k})
    \right\|_2^2.
    \label{eq:pwm-objective}
\end{aligned}
\end{equation}
This mean-matching anchor preserves the pretrained transition
dynamics while the group-relative advantage moves the adapter
toward better continuations for the target scene.
We use four Euler denoising steps, four PPO passes per rollout
group, $\lambda=0.1$, a learning rate of $10^{-4}$, and SDE
temperature $\beta=0.5$.
For the GRPO instantiation, every scene is trained for eight
epochs with the same configuration.

\subsection{Per-scene Customization via DiffusionNFT}
\label{sec:method-nft}

We also instantiate PWM with
DiffusionNFT~\citep{zheng2026diffusionnft}, which incorporates
reward feedback through a forward-process flow-matching
objective.
This instantiation uses the same frozen Yume-5B backbone,
LoRA capacity, six native support windows, and unified
reward as the GRPO instantiation.

\paragraph*{Online sampling and reward normalization.}
For each support window, we sample $G=4$ continuations with
independent initial latent noise and compute their rewards
using the same score-on-mean procedure.
Let $R_i$ be the scalar reward from Eq.~\eqref{eq:reward},
and let $\overline{R}$ be its mean within the current window's
rollout group.
We convert each reward into a bounded weight
\begin{equation}
    q_i =
    \frac{1}{2}
    +\frac{1}{2}
    \operatorname{clip}\left(
        \frac{R_i-\overline{R}}{Z_s},
        -1,1
    \right),
    \label{eq:nft-reward}
\end{equation}
where $Z_s$ is the running standard deviation of rollout
rewards collected during adaptation of scene $s$, with
a lower bound of $10^{-3}$.

\paragraph*{Reward-guided flow matching.}
Let $z_i^0$ denote the generated latent segment used for
reward measurement.
We sample a noise level $\sigma$ from the four levels of
the generation schedule and draw fresh Gaussian noise:
\begin{equation}
\begin{gathered}
    z_i^\sigma=(1-\sigma)z_i^0+\sigma\epsilon,
    \qquad \epsilon\sim\mathcal{N}(0,I),\\
    v_i^\star=\epsilon-z_i^0.
    \label{eq:nft-forward}
\end{gathered}
\end{equation}
The visual history and action condition are kept fixed.
Let $v_\theta$ be the current velocity prediction at
$(z_i^\sigma,\sigma,c_t)$ and $v_{\mathrm{old}}$ the
prediction cached before the update.
DiffusionNFT defines
\begin{equation}
\begin{aligned}
    v_i^+ &=
    (1-\beta_{\mathrm{NFT}})v_{\mathrm{old}}
    +\beta_{\mathrm{NFT}}v_\theta,\\
    v_i^- &=
    (1+\beta_{\mathrm{NFT}})v_{\mathrm{old}}
    -\beta_{\mathrm{NFT}}v_\theta,
    \label{eq:nft-velocities}
\end{aligned}
\end{equation}
and minimizes
\begin{equation}
\begin{aligned}
    \mathcal{L}_{\mathrm{NFT}}
    =\mathbb{E}_{i,\sigma,\epsilon}\Big[
    &q_i\,\operatorname{MSE}(v_i^+,v_i^\star)\\
    &+(1-q_i)\,
    \operatorname{MSE}(v_i^-,v_i^\star)
    \Big].
    \label{eq:nft-objective}
\end{aligned}
\end{equation}
Here, $\operatorname{MSE}$ averages over latent elements,
and gradients are taken only through $v_\theta$.
The reward weights determine the contribution of the positive
and negative branches to the adapter update.

We collect $24$ rollouts across the six support windows and
draw two noise-level/noise pairs per rollout.
We perform four optimization passes over this collected batch,
accumulating gradients across windows.
The cached behavior predictions remain fixed during these passes.
We use $\beta_{\mathrm{NFT}}=0.5$, a learning rate of
$5\times10^{-5}$, and eight training epochs.
This instantiation uses the current behavior-policy snapshot
without an EMA teacher or an additional base-policy anchor.

\subsection{A Unified Reward Across Domains}
\label{sec:method-reward}

Both the GRPO and DiffusionNFT instantiations of PWM use
one reward for all scenes and first-person domains:
\begin{equation}
    R
    =0.50\,R_{\mathrm{gt}}
    +0.30\,R_{\mathrm{lpips}}
    +0.10\,R_{\mathrm{cross}}
    +0.10\,R_{\mathrm{dyn}}.
    \label{eq:reward}
\end{equation}
The four terms give the optimizer distinct responsibilities.
$R_{\mathrm{gt}}$ measures spatially aligned DINO feature
agreement between the generated segment and the real
continuation of the support window.
$R_{\mathrm{lpips}}$ is one minus their
LPIPS~\citep{zhang2018lpips} distance and therefore preserves
perceptual appearance.
Both terms are computed per frame and aggregated as
$0.7$ times the temporal mean plus $0.3$ times the minimum,
so a short failure cannot be hidden by otherwise strong frames.
$R_{\mathrm{cross}}$ measures DINO feature agreement between
the first generated frame and the window's ground-truth first frame.
Finally, $R_{\mathrm{dyn}}$ rewards agreement between generated
and real optical-flow magnitudes, so appearance fidelity cannot
be obtained by suppressing motion.
Together, the first two terms specify what the scene should
look like, while the latter two constrain how a valid
continuation should evolve.

All reward references are drawn exclusively from the support
trajectory. We do not tune reward weights per scene or per
domain, and the evaluation metric is separate from this
training reward (Section~\ref{sec:benchmark}).
This separation makes PWM a single customization procedure
rather than a collection of scene-specific reward recipes.

GRPO combines component-wise normalized advantages through
Eq.~\eqref{eq:component-advantage}, while DiffusionNFT uses the
combined scalar reward through Eq.~\eqref{eq:nft-reward}.

\subsection{Checkpoint Soup}
\label{sec:method-soup}

Small-data policy optimization can visit useful but different
adapter states across epochs.
Selecting one of them per scene would introduce an additional
selection rule and could couple the delivered model to a noisy
validation estimate.
For both GRPO and DiffusionNFT, PWM instead saves the adapter
at every epoch and averages all eight checkpoints element-wise:
\begin{equation}
    \overline{\Delta\theta}_s
    =
    \frac{1}{8}\sum_{e=1}^{8}\Delta\theta_s^{(e)}.
    \label{eq:soup}
\end{equation}
Here, averaging is applied to the corresponding stored LoRA
factor tensors.
This checkpoint soup~\citep{wortsman2022modelsoups} is the
final scene model.
Because factor-wise averaging preserves the adapter's rank
and architecture, PWM still deploys one compact adapter and
incurs no additional inference cost.
The delivery rule is fixed across scenes and does not inspect
the future test trajectory.

\section{PWM-Bench}
\label{sec:benchmark}

\begin{figure}[!tb]
    \centering
    \includegraphics[width=\textwidth]{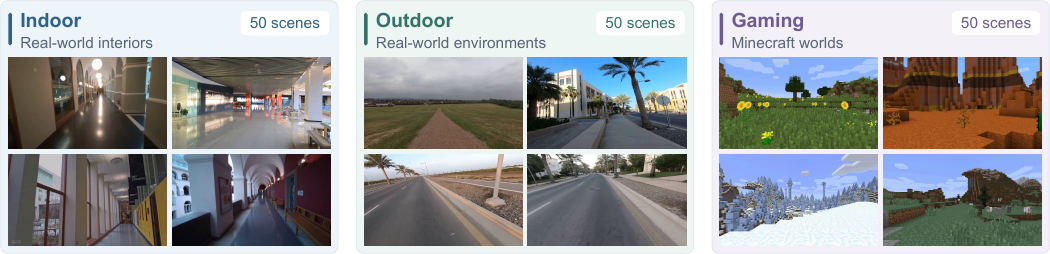}
    \caption{\textbf{Training scenes in PWM-Bench.}
    Each domain contains 50 scenes; four representative training scenes
    are shown per domain. Each image is the full, unretouched frame at
    zero-based index 40 of an 80-frame support clip.}
    \label{fig:pwm-bench-training}
\end{figure}

Customization promises that the model reproduces the persistent
identity of \emph{this} scene --- its layout, materials, and sense
of place --- on a continuation it has never observed.
Our benchmark evaluates this promise directly: each generated
rollout is scored against the held-out ground-truth continuation
of its own scene.

Figure~\ref{fig:pwm-bench-training} shows representative training-support
frames from the three domains of PWM-Bench.

\subsection{Customization-Fidelity Score}
\label{sec:bench-metric}

For a generated rollout $g_{1:T}$ and the ground-truth test
segment $x_{1:T}$ ($T{=}145$ frames), the score is
\begin{align}
    S
    &= D_{\mathrm{pair}} + 0.5\,D_{\mathrm{hall}},
    \label{eq:sgt}\\
    D_{\mathrm{pair}}
    &= \frac{1}{T}\sum_{t=1}^{T}
    \Bigl(
        1-\cos\bigl(\phi(g_t),\phi(x_t)\bigr)
    \Bigr),
    \label{eq:dpair}\\
    D_{\mathrm{hall}}
    &= \frac{1}{T}\sum_{t=1}^{T}
    \mathbf{1}\bigl[
        g_t \text{ contains an entity class absent from } x_1
    \bigr],
    \label{eq:dhall}
\end{align}
where $\phi(\cdot)$ is the DINOv2 class-token
embedding~\citep{oquab2023dinov2}.
Lower $S$ indicates higher fidelity to the scene's true
continuation.

$D_{\mathrm{pair}}$ measures appearance fidelity as the
average feature distance between generated and ground-truth
frames at the same timestep, in the DINO feature space
commonly used to measure fidelity in personalized
generation~\citep{ruiz2023dreambooth}.
We use a time-aligned comparison over the full horizon,
rather than a distributional one, because distributional
metrics can reward plausible but generic continuations;
matching the specific scene as it unfolds is precisely
what customization is expected to deliver.

$D_{\mathrm{hall}}$ measures the fraction of generated
frames containing an entity class
(person, vehicle, or animal, detected by zero-shot
CLIP~\citep{radford2021clip})
that is absent from the first frame of the ground-truth
test segment.
It serves as a proxy for unexpected entity introduction
relative to the initial scene.
The detection rules and thresholds are specified in
Appendix~\ref{app:score-implementation}.

Both PWM instantiations and Yume are scored by the same
pipeline, with identical captions, initial context,
and generation horizon.
For each instantiation $m$, we report the paired difference
\begin{equation}
    \Delta_m = S(m)-S(\mathrm{Yume}),
\end{equation}
where $m$ denotes PWM (GRPO) or PWM (DiffusionNFT).
A negative $\Delta_m$ indicates improved customization
over Yume.

\subsection{Scoring Protocol}
\label{sec:bench-scoring}

Each scene is evaluated with three random seeds,
shared across Yume and both PWM instantiations,
under deterministic inference.
For each instantiation, a scene counts as improved
if its mean paired difference against Yume over
the three seeds is negative.
We report improvement rates and mean and maximum
scene-level relative gains by domain.

Adaptation uses only the support windows, and evaluation
uses only the held-out test segment, so no evaluated
frame is seen during training.
Both PWM instantiations are evaluated on the same
fixed set of scenes.

\section{Experimental Setup}
\label{sec:setup}

\subsection{Cross-domain Evaluation Suite}
\label{sec:setup-suite}

We evaluate 150 independent customization tasks across three
first-person domains: 50 outdoor walking scenes, 50 indoor walking
scenes, and 50 Minecraft scenes (Table~\ref{tab:setup-domains}).
Each task has its own support trajectory, scene adapter, and
held-out test trajectory.

\begin{table}[!htbp]
    \centering
    \small
    \setlength{\tabcolsep}{3pt}
    \caption{Composition of the cross-domain evaluation suite.
    Each domain contains 50 independent per-scene customization tasks.}
    \label{tab:setup-domains}
    \begin{tabular}{@{}lc
        p{0.27\columnwidth}
        p{0.28\columnwidth}@{}}
        \toprule
        Domain & Tasks & Video source & Control source \\
        \midrule
        Outdoor & 50 & EgoVid-5M, YouTube & VIO / visual motion \\
        Indoor & 50 & EgoVid-5M, InCrowd-VI, YouTube & VIO / visual motion \\
        Gaming  & 50 & MineDojo & Scripted-agent logs \\
        \bottomrule
    \end{tabular}
\end{table}

Real-world videos come from the visual--inertial odometry (VIO)
annotated subset of EgoVid-5M~\citep{egovid5m}, YouTube, and,
for Indoor, InCrowd-VI~\citep{bamdad2024incrowd}.
YouTube camera motion is estimated from video and mapped to the
shared control representation.

Following GameFactory's~\citep{gamefactory} automated collection
paradigm, our scripted agents record Minecraft videos and synchronized
locomotion and camera controls in MineDojo~\citep{minedojo}.

\subsection{Per-scene Task Construction}
\label{sec:setup-tasks}

Each task contains an 80-frame support segment and a temporally later,
non-overlapping 145-frame test segment from the same scene.
PWM adapts independently on each support segment and evaluates the
resulting adapter on its held-out continuation.

We use a shared control interface across all three domains. Each video
window is conditioned on translation commands (W/A/S/D), a
camera-direction token, and three continuous motion magnitudes. For
EgoVid-5M, these annotations are obtained from the associated VIO
trajectory. For YouTube videos, they are derived from estimated camera
motion. For Minecraft, they are recorded directly from the scripted
agents.

The same source-side quality criteria are applied across domains. The
retained tasks provide continuous and non-degenerate camera motion,
consistent scene identity across the support and test segments, and
agreement between the control annotations and observed visual motion.

\subsection{Controlled Comparisons}
\label{sec:setup-comparisons}

We compare PWM with native Yume-5B using the same 29-frame
autoregressive generation process, initial visual context, controls,
decoding configuration, and rollout horizon.
Native Yume uses the same inference pipeline with the scene adapter
disabled, isolating the effect of per-scene adaptation.

We additionally compare PWM with a matched SFT-LoRA baseline in the
reinforcement-learning ablation. SFT-LoRA uses the same frozen Yume-5B
backbone, LoRA capacity, support data, control annotations, optimization
budget, and evaluation protocol as PWM. This comparison isolates the
contribution of online GRPO beyond scene-specific parameter updates alone.

\begin{figure}[t]
    \centering
    \includegraphics[width=\textwidth]{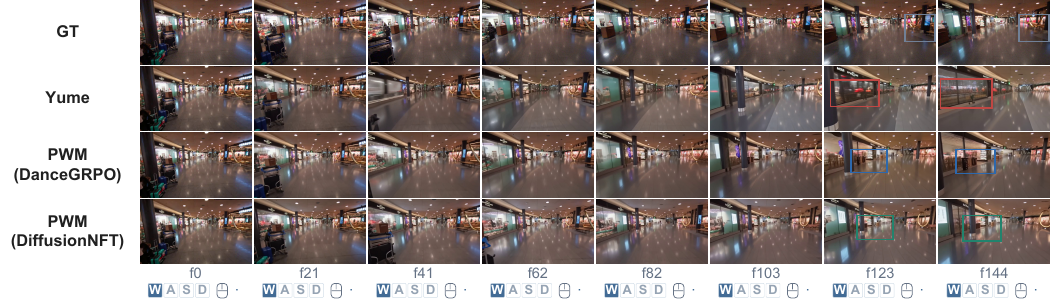}
    \caption{\textbf{Scene appearance in a shopping-mall continuation.}
    Rows share the controls; columns show frames
    0, 21, 41, 62, 82, 103, 123, and 144.
    Yume replaces shop displays with window-like structures;
    both PWM variants retain shop displays.
    Boxes mark selected regions; GT denotes ground truth.}
    \label{fig:scene-appearance}
\end{figure}

\subsection{Shared Training and Evaluation Configuration}
\label{sec:setup-configuration}

All videos are processed at a resolution of $832\times480$. Each scene is
adapted independently with a rank-8 LoRA using $\alpha=16$, corresponding
to 20.2M trainable parameters. We instantiate PWM with GRPO and DiffusionNFT, as described
in Sections~\ref{sec:method-grpo} and~\ref{sec:method-nft}.
Both instantiations use the same adapter capacity and
unified reward in Eq.~\eqref{eq:reward}, with checkpoint
averaging described in Section~\ref{sec:method-soup}.
Within each instantiation, the optimization configuration
and checkpoint-averaging rule are fixed across all 150 tasks.

Evaluation follows Section~\ref{sec:bench-scoring}. Each scene is evaluated
with three paired random seeds, $\{42,123,777\}$, shared between PWM and
native Yume under deterministic inference. Both methods generate the full
145-frame test trajectory. Seed-level paired differences are first averaged
within each scene, and every scene contributes once to its domain and to the
overall benchmark.

\section{Experiments}
\label{sec:experiments}

\begin{wrapfigure}{r}{0.5\textwidth}
\vspace{-2em}
    \centering
    \includegraphics[width=\linewidth]{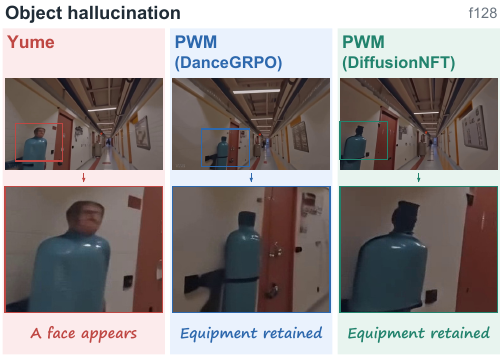}
    \caption{\textbf{Object hallucination indoors.}
        Yume adds a face to blue equipment; both PWM variants retain
        its appearance. Methods share the seed and controls (frame 128).
        Lower panels enlarge equally sized boxed crops.}
    \label{fig:object-hallucination}
    \vspace{-3em}
\end{wrapfigure}

We evaluate PWM along three dimensions: customization gains over
native Yume, the contribution of reinforcement learning beyond
supervised adaptation, and visual quality after customization.
The evaluation covers 150 scenes across Indoor, Outdoor, and
Gaming, with three paired generation seeds per scene.
All customization scores are computed over the complete
145-frame held-out continuation, following
Section~\ref{sec:setup}.

\subsection{Scene Customization across Domains}
\label{sec:exp-main}

\begin{figure}[t]
    \centering
    \includegraphics[width=\textwidth]{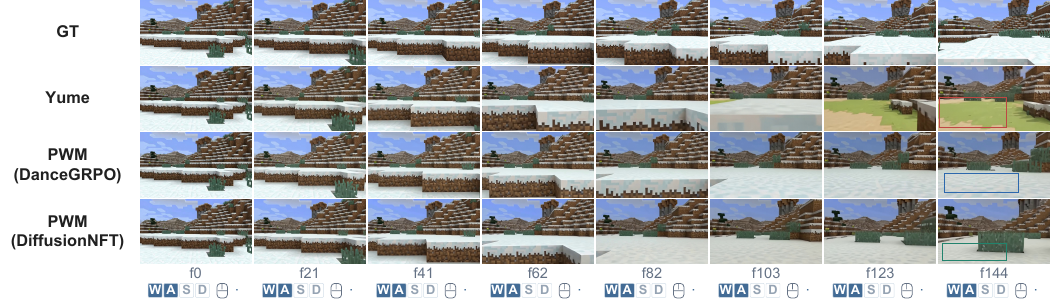}
    \caption{\textbf{Snow-cover retention in Gaming.}
    Both PWM variants retain snow-covered ground.
    Rows share controls and use the frame indices in
    Figure~\ref{fig:scene-appearance}.
    GT denotes ground truth; boxes mark selected regions.}
    \label{fig:snow-comparison}
\end{figure}

\begin{figure}[t]
    \centering
    \includegraphics[width=\textwidth]{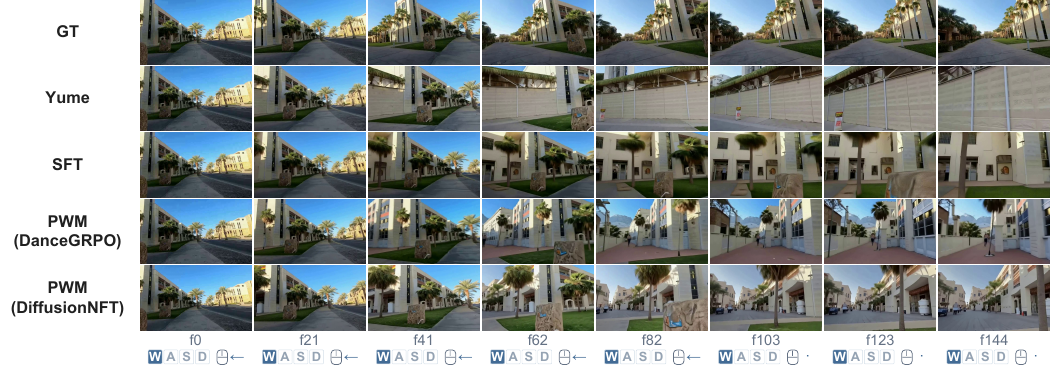}
    \caption{\textbf{Scene adaptation outdoors.}
    Methods share the seed, controls, and eight uniformly sampled frame
    indices. Yume shows a nearby wall and SFT close-ups of buildings
    and palms; both PWM variants retain visible passage space.}
    \label{fig:rl-sft-qualitative}
\end{figure}

\paragraph*{Scene appearance and content retention.}
Figure~\ref{fig:scene-appearance} illustrates scene customization in a shopping-mall continuation. Both PWM instantiations retain shop displays, while Yume replaces them with window-like structures in later frames.

\paragraph*{Measuring customization benefit.}
We quantify customization benefit as the relative reduction
in the score $S$ defined in Eq.~\eqref{eq:sgt}.
For scene $s$ and method $m$, let $\bar{S}_s^{m}$ denote
the score averaged over the three generation seeds.
The scene-level gain, expressed as a percentage, is
\begin{equation}
    g_s(m)
    =
    100\,
    \frac{
        \bar{S}_s^{\mathrm{Native}}-\bar{S}_s^{m}
    }{
        \bar{S}_s^{\mathrm{Native}}
    }.
    \label{eq:exp-relative-gain}
\end{equation}
Positive values indicate improved customization over Native.
We report the mean gain across all evaluated scenes,
the maximum scene-level gain, and the fraction of scenes
with $g_s(m)>0$.
Every scene receives equal weight, and negative gains
are included in the average.
Both the mean and maximum are computed after averaging
the three seeds within each scene.

\clearpage

\begin{wraptable}{r}{0.65\textwidth}
\vspace{-1em}
    \centering
    \small
    \setlength{\tabcolsep}{3pt}
    \caption{
        \textbf{PWM improves customization across all three domains.}
        Both optimization instantiations are compared with native Yume.
        Gains are relative reductions in $S$.
        Mean gain includes all evaluated scenes.
        Maximum gain and improved-scene counts are computed
        after averaging three seeds within each scene.
    }
    \label{tab:main-results}
    \begin{tabular}{@{}lrrr@{}}
        \toprule
        Method
        & \shortstack{Mean gain\\(\%) $\uparrow$}
        & \shortstack{Max.\ gain\\(\%) $\uparrow$}
        & \shortstack{Improved\\scenes $\uparrow$} \\
        \midrule
        \multicolumn{4}{@{}l}{\textit{Indoor}} \\
        PWM (DanceGRPO)
        & 5.14 & 49.00 & 36/50 (72.0\%) \\
        PWM (DiffusionNFT)
        & 7.04 & 58.54 & 34/50 (68.0\%) \\
        \midrule
        \multicolumn{4}{@{}l}{\textit{Outdoor}} \\
        PWM (DanceGRPO)
        & 6.11 & 75.11 & 34/50 (68.0\%) \\
        PWM (DiffusionNFT)
        & 2.27 & 46.38 & 30/50 (60.0\%) \\
        \midrule
        \multicolumn{4}{@{}l}{\textit{Gaming}} \\
        PWM (DanceGRPO)
        & 7.91 & 28.66 & 37/50 (74.0\%) \\
        PWM (DiffusionNFT)
        & 6.35 & 38.76 & 34/50 (68.0\%) \\
        \midrule
        \multicolumn{4}{@{}l}{\textit{Overall}} \\
        PWM (DanceGRPO)
        & 6.39 & 75.11 & 107/150 (71.3\%) \\
        PWM (DiffusionNFT)
        & 5.22 & 58.54 & 98/150 (65.3\%) \\
        \bottomrule
    \end{tabular}
    \vspace{-2em}
\end{wraptable}

\paragraph*{Positive gains across real and rendered environments.}
The GRPO and DiffusionNFT instantiations of PWM improve
customization in 107 and 98 of the 150 evaluated scenes,
respectively, achieving mean relative gains of
$6.39\%$ and $5.22\%$ over native Yume
(Table~\ref{tab:main-results}).
The corresponding improvement rates are
$71.3\%$ and $65.3\%$.
Both instantiations achieve positive mean gains across
Indoor, Outdoor, and Gaming.
These results support the effectiveness of per-scene
customization across real and rendered environments
using either online reward-guided optimization method.

\paragraph*{Benefits beyond the strongest examples.}
For the GRPO instantiation, the largest scene-level gains reach $49.00\%$ in Indoor,
$75.11\%$ in Outdoor, and $28.66\%$ in Gaming.
The median gains are also positive in all three domains:
$5.18\%$, $5.82\%$, and $8.18\%$, respectively.
Together, the average gain, median gain, and improvement
frequency show that the observed customization benefit
extends beyond a few highly successful examples.

\begin{wraptable}{r}{0.58\textwidth}
\vspace{-2em}
    \centering
    \small
    \setlength{\tabcolsep}{3pt}
    \caption{
        \textbf{Both PWM instantiations achieve higher mean
        customization gains than SFT across all three domains.}
        All methods are compared with native Yume.
        Improved scenes have a lower three-seed mean $S$
        than Native.
        Bold indicates the best result within each domain.
    }
    \label{tab:rl-sft}
    \begin{tabular}{@{}lrr@{}}
        \toprule
        Method
        & \shortstack{Mean gain\\(\%) $\uparrow$}
        & \shortstack{Improved\\scenes $\uparrow$} \\
        \midrule
        \multicolumn{3}{@{}l}{\textit{Indoor}} \\
        SFT
        & $-16.39$ & 27/50 (54.0\%) \\
        PWM (DanceGRPO)
        & $+5.14$ & \textbf{36/50 (72.0\%)} \\
        PWM (DiffusionNFT)
        & $\mathbf{+7.04}$ & 34/50 (68.0\%) \\
        \midrule
        \multicolumn{3}{@{}l}{\textit{Outdoor}} \\
        SFT
        & $-14.86$ & 26/50 (52.0\%) \\
        PWM (DanceGRPO)
        & $\mathbf{+6.11}$ & \textbf{34/50 (68.0\%)} \\
        PWM (DiffusionNFT)
        & $+2.27$ & 30/50 (60.0\%) \\
        \midrule
        \multicolumn{3}{@{}l}{\textit{Gaming}} \\
        SFT
        & $+6.08$ & 33/50 (66.0\%) \\
        PWM (DanceGRPO)
        & $\mathbf{+7.91}$ & \textbf{37/50 (74.0\%)} \\
        PWM (DiffusionNFT)
        & $+6.35$ & 34/50 (68.0\%) \\
        \midrule
        \multicolumn{3}{@{}l}{\textit{Overall}} \\
        SFT
        & $-8.39$ & 86/150 (57.3\%) \\
        PWM (DanceGRPO)
        & $\mathbf{+6.39}$ & \textbf{107/150 (71.3\%)} \\
        PWM (DiffusionNFT)
        & $+5.22$ & 98/150 (65.3\%) \\
        \bottomrule
    \end{tabular}
    \vspace{-5em}
\end{wraptable}

\paragraph*{Snow-cover retention in Gaming.}
Figure~\ref{fig:snow-comparison} complements the indoor example
with a rendered environment: both PWM instantiations retain
snow-covered ground in the selected continuations.

\paragraph*{Object hallucination.}
Figure~\ref{fig:object-hallucination} shows a local object
hallucination: a human face appears on the blue equipment
in Yume, while both PWM instantiations retain its
equipment appearance.

\subsection{The Benefit of Reinforcement Learning}
\label{sec:exp-rlsft}

\paragraph*{Comparison with matched supervised adaptation.}
We assess online RL against matched SFT-LoRA under the settings
in Section~\ref{sec:setup-comparisons}.
Customization gains and improved-scene counts are measured
against the same native Yume baseline.

Figure~\ref{fig:rl-sft-qualitative} provides a qualitative
comparison of Native Yume, SFT, and both PWM instantiations.
In this outdoor example, both PWM instantiations retain
visible passage space as generation progresses.

\paragraph*{Higher average customization benefit.}
Both the DanceGRPO and DiffusionNFT instantiations of PWM
achieve higher mean gains than SFT in every domain
(Table~\ref{tab:rl-sft}).
In Indoor and Outdoor, both instantiations achieve
positive mean gains, compared with negative mean gains
for SFT.
In Gaming, the mean gain increases from $6.08\%$ with
SFT to $7.91\%$ with DanceGRPO and $6.35\%$ with DiffusionNFT.
Across all 150 scenes, the mean gain rises from
$-8.39\%$ with SFT to $+6.39\%$ with DanceGRPO and
$+5.22\%$ with DiffusionNFT.

\begin{wrapfigure}{R}{0.5\textwidth}
\vspace{-2em}
    \centering
    \includegraphics[width=\linewidth]{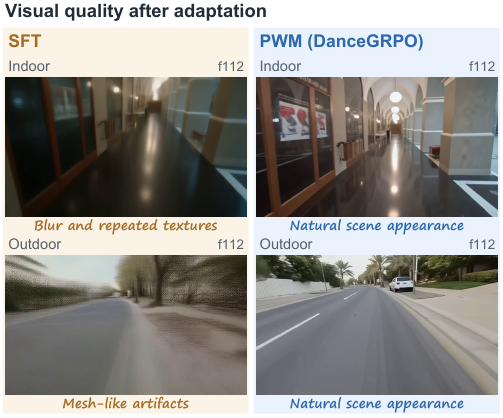}
    \caption{
        \textbf{Preserving visual quality after adaptation.}
        SFT exhibits blur and repetitive textures indoors
        and mesh-like artifacts outdoors, while
        PWM (DanceGRPO) retains natural-looking scene appearance.
        Each pair uses the same seed and control sequence
        at frame 112. All panels show complete,
        unretouched frames.
    }
    \label{fig:sft-visual-quality}
    \vspace{-3em}
\end{wrapfigure}

\paragraph*{Scene-level improvement coverage.}
Across all 150 scenes, the DanceGRPO and DiffusionNFT
instantiations of PWM improve over Native in 107 and
98 scenes, respectively, compared with 86 scenes for SFT.
The corresponding improvement rates are $71.3\%$,
$65.3\%$, and $57.3\%$.
Together with the higher mean gains across all three
domains, these results show that both PWM instantiations
deliver stronger average customization and benefit
a larger fraction of the overall evaluation suite
than SFT.

The matched SFT comparison examines how support observations
are converted into a scene-specific generation policy.
SFT optimizes a supervised diffusion objective on the observed
support windows.
PWM instead samples continuations from the current policy and
uses their relative rewards to guide adaptation.
The learning signal therefore depends on what the adapted model
actually generates under the support conditions.

\subsection{Preserving Visual Quality}
\label{sec:exp-quality}

Figure~\ref{fig:sft-visual-quality} illustrates this contrast:
SFT exhibits visible blur and repetitive texture artifacts,
while PWM (DanceGRPO) retains natural-looking scene appearance
in the selected examples.

\begin{wraptable}{R}{0.65\textwidth}
\vspace{-2em}
    \centering
    \small
    \setlength{\tabcolsep}{3pt}
    \caption{
        \textbf{Both PWM variants achieve better mean
        image-quality scores than SFT in every domain.}
        Native Yume provides the pretrained reference.
        Bold indicates the best mean within each domain.
        MUSIQ uses the SPAQ-trained checkpoint.
    }
    \label{tab:visual-quality}
    \begin{tabular}{@{}lrrr@{}}
        \toprule
        Method
        & MUSIQ $\uparrow$
        & NIQE $\downarrow$
        & BRISQUE $\downarrow$ \\
        \midrule
        \multicolumn{4}{@{}l}{\textit{Indoor}} \\
        Yume
        & 58.93 & 5.43 & 44.46 \\
        SFT
        & 47.26 & 5.93 & 45.73 \\
        PWM (DanceGRPO)
        & \textbf{59.06} & \textbf{5.40} & \textbf{44.28} \\
        PWM (DiffusionNFT)
        & 57.94 & 5.49 & 44.79 \\
        \midrule
        \multicolumn{4}{@{}l}{\textit{Outdoor}} \\
        Yume
        & 71.33 & \textbf{3.65} & \textbf{22.66} \\
        SFT
        & 63.06 & 4.64 & 33.58 \\
        PWM (DanceGRPO)
        & 71.51 & 3.68 & 23.24 \\
        PWM (DiffusionNFT)
        & \textbf{71.58} & 3.67 & 22.74 \\
        \midrule
        \multicolumn{4}{@{}l}{\textit{Gaming}} \\
        Yume
        & 74.98 & 3.81 & 28.82 \\
        SFT
        & 73.23 & 5.90 & 29.73 \\
        PWM (DanceGRPO)
        & 75.16 & 3.82 & 28.11 \\
        PWM (DiffusionNFT)
        & \textbf{75.40} & \textbf{3.76} & \textbf{27.99} \\
        \bottomrule
    \end{tabular}
    \vspace{-2em}
\end{wraptable}

\paragraph*{Quality assessment alongside customization.}
We complement the customization score with three
no-reference image-quality metrics:
MUSIQ-SPAQ~\citep{ke2021musiq},
NIQE~\citep{mittal2013niqe}, and
BRISQUE~\citep{mittal2012brisque}.
These metrics characterize frame-level visual quality,
while $S$ evaluates scene-specific customization.
MUSIQ uses all frames; NIQE and BRISQUE use 19
uniformly sampled frames, including the first and last.
Scores are averaged over frames, seeds, and scenes,
in that order.
Native Yume provides a reference for the pretrained
model's visual quality.

\paragraph*{Quality close to Native after customization.}
PWM outperforms SFT on the domain means of all three
reported image-quality metrics
(Table~\ref{tab:visual-quality}).
In Indoor, MUSIQ increases from $47.26$ with SFT
to $59.06$ with PWM, compared with $58.93$ for Native.
In Outdoor, BRISQUE improves from $33.58$ to $23.24$,
close to Native's $22.66$.
In Gaming, NIQE improves from $5.90$ to $3.82$,
compared with $3.81$ for Native.
Across individual scenes, MUSIQ favors PWM over SFT
in 123 of 150 cases, and NIQE does so in 132 cases.

Scene identity and image quality provide complementary evidence.
The customization score evaluates agreement with the target scene,
while the image-quality metrics characterize frame-level appearance.
Together, Tables~\ref{tab:main-results} and~\ref{tab:visual-quality}
support the intended benefit of PWM: stronger fidelity to the target
scene with frame-level quality close to native Yume.

\subsection{Training Dynamics}
\label{sec:training-dynamics}
\begin{figure}[H]
    \centering
    \vspace{-6pt}
    \includegraphics[width=\textwidth]{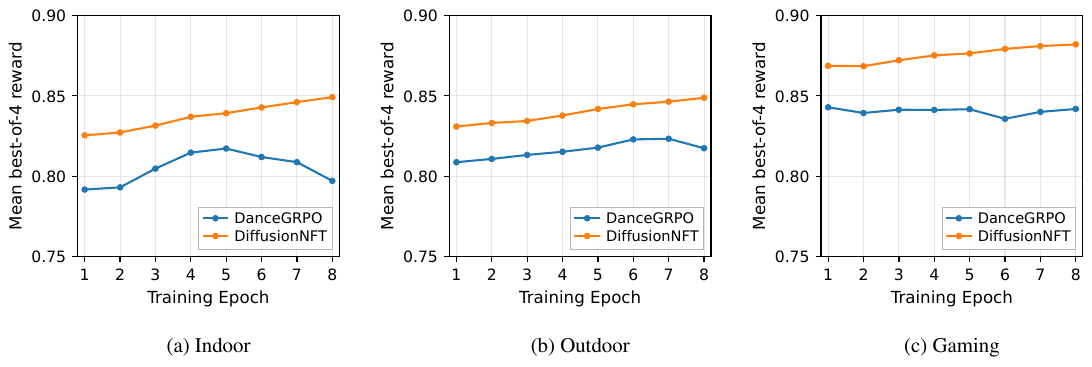}
    \par\vspace{-10pt}
    \caption{\textbf{Training reward trajectories.}
    Each panel averages 50 scenes.
    Per-scene rewards average the best of four rollouts
    over six support windows in each epoch.
    Rewards are recorded under each method's
    training sampling procedure.}
    \label{fig:training-reward}
\end{figure}

Figure~\ref{fig:training-reward} summarizes the training
reward trajectories of PWM with DanceGRPO and DiffusionNFT
across Indoor, Outdoor, and Gaming.
The curves report rewards collected under each method's
training sampling procedure, complementing the held-out
customization results in Table~\ref{tab:main-results}.

\section{Conclusion and Future Work}
\label{sec:conclusion}

We introduced PWM, a framework for customizing a pretrained
interactive world model to an individual scene through
online reinforcement learning.
PWM turns a short support trajectory into a compact LoRA
adapter using native-window generation, a shared reward,
and a fixed checkpoint-averaging rule.
We also introduced PWM-Bench to evaluate per-scene
customization across Indoor, Outdoor, and Gaming.
Across the evaluated suite, PWM improves customization over
native Yume in 107 of 150 scenes.
The matched SFT comparison further demonstrates higher mean
customization gains and broader scene-level benefits, alongside
visual quality close to the pretrained model.

Future work will extend this approach to longer interaction
horizons and richer support trajectories, enabling scene
customization over more extensive exploration.

\subsubsection*{Reproducibility Statement}

Section~\ref{sec:setup} specifies task construction and the
shared experimental protocol.
Appendices~\ref{app:implementation}--\ref{app:evaluation}
provide optimization settings, reward definitions, and scoring details.
Evaluation uses three paired seeds on held-out continuations,
with no test frames used for adaptation or checkpoint construction.

\subsubsection*{AI Use Declaration}
OpenAI ChatGPT was used to assist with language polishing
and improving the clarity of the manuscript.
The authors take full responsibility for the final content.

\clearpage
\bibliography{egbibsample}
\bibliographystyle{iclr2027_conference}

\clearpage

\appendix
\section{Implementation Details}
\label{app:implementation}

This appendix describes the DanceGRPO implementation.
DiffusionNFT follows Section~\ref{sec:method-nft}.
The checkpoint averaging procedure described below
applies to both instantiations.
The four reward components are specified in
Appendix~\ref{app:rewards}.

\subsection{Adaptation Configuration}
\label{app:algorithm}

For each scene, we train one LoRA adapter on the frozen Yume-5B
backbone.
The adapter is shared across six overlapping support windows of
$29$ frames, starting at zero-based indices
$[0,9,18,27,36,45]$.
Each window is conditioned on its own ground-truth first frame
and corresponding control caption.
Windows are processed in this order, with adapter updates
carried forward within the scene.

For each window, the current policy generates a group of $G=4$
trajectories using $K=4$ stochastic denoising steps.
We compute the trajectory advantages and perform four optimization
passes on the collected group.
Each pass accumulates gradients over all $GK$ transitions before
taking one optimizer step.
We train for eight epochs and save the adapter after each complete
pass through the six windows.
This gives $48$ rollout groups and $192$ optimizer updates per scene.

Table~\ref{tab:app-hyperparameters} lists the shared configuration.
Backbone forward passes use bfloat16, and trainable LoRA parameters
are stored in float32.
Reward networks are frozen and evaluated without gradient
propagation through them.

\begin{table}[!htbp]
    \centering
    \small
    \setlength{\tabcolsep}{3pt}
    \caption{
        PWM (DanceGRPO) adaptation configuration,
        fixed across scenes. Reward weights follow the order
        $(\mathrm{gt},\mathrm{lpips},\mathrm{cross},\mathrm{dyn})$.
    }
    \label{tab:app-hyperparameters}
    \begin{tabular}{@{}
        >{\raggedright\arraybackslash}p{0.47\columnwidth}
        >{\raggedright\arraybackslash}p{0.47\columnwidth}@{}}
        \toprule
        Setting & Value \\
        \midrule
        Frozen backbone & Yume-5B \\
        LoRA rank / scaling $\alpha$ & $8$ / $16$ \\
        LoRA target modules
        & \texttt{q, k, v, o,}
          \newline \texttt{ffn.0, ffn.2} \\
        LoRA dropout / trainable biases & $0$ / none \\
        Trainable parameters & Approximately $20.2$M \\
        Window length & $29$ frames \\
        Windows per epoch & $6$ \\
        Group size / denoising steps & $4$ / $4$ \\
        Epochs / passes per group & $8$ / $4$ \\
        Optimizer & AdamW \\
        Learning rate / weight decay & $10^{-4}$ / $10^{-2}$ \\
        Clipping $\varepsilon_{\mathrm{clip}}$ & $0.2$ \\
        Maximum gradient norm & $1.0$ \\
        Base-anchor coefficient $\lambda$ & $0.1$ \\
        Stochastic coefficient $\beta$ & $0.5$ \\
        Reward weights & $0.50, 0.30, 0.10, 0.10$ \\
        Checkpoint averaging & Uniform over all eight epochs \\
        \bottomrule
    \end{tabular}
\end{table}

\subsection{Trajectory Sampling and Reward Measurement}
\label{app:sampling}

Let $z_k$ denote the generated latent variables at denoising
step $k$, $c$ the conditioning information, and $\sigma_k$
the noise schedule.
The stochastic Euler transition is
\begin{align}
\mu_\theta(z_k,c)
&=z_k+(\sigma_{k-1}-\sigma_k)
  v_\theta(z_k,c,\sigma_k),\\
\eta_k
&=\beta\sqrt{|\sigma_{k-1}-\sigma_k|}
  \max(\sigma_k,10^{-4})+10^{-4},\\
z_{k-1}
&=\mu_\theta(z_k,c)+\eta_k\epsilon_k,
\qquad \epsilon_k\sim\mathcal{N}(0,I).
\end{align}
Trajectories within a group share their initial latent noise
and use independent transition noise.
We record the sampled transitions and their behavior-policy
log densities for subsequent optimization.

For reward measurement, we decode the conditional mean of the
final transition.
This excludes the final noise injection from the scored video
while retaining the effects of earlier stochastic transitions.
Policy optimization uses the recorded stochastic transitions.

\subsection{Advantage Computation and Policy Updates}
\label{app:advantages}

Each reward component is normalized within the rollout group
before the component advantages are combined.
For component $m$, let
\begin{equation}
\mu_m=\frac{1}{G}\sum_{i=1}^{G}R_{i,m},
\qquad
\sigma_m=
\sqrt{\frac{1}{G}\sum_{i=1}^{G}(R_{i,m}-\mu_m)^2}.
\end{equation}
Using the fixed component weights $\alpha_m$, we compute
\begin{equation}
A_i=\sum_m\alpha_m\,
\mathbf{1}[\sigma_m\geq10^{-3}]
\frac{R_{i,m}-\mu_m}{\sigma_m+10^{-6}}.
\label{eq:app-advantage}
\end{equation}
Components with standard deviation below $10^{-3}$ contribute
zero to that group's advantage.
The same $A_i$ is assigned to all denoising steps of trajectory $i$.

For a generation region containing $d$ scalar latent variables,
we use the mean Gaussian log density
\begin{equation}
\ell_\theta(z_{k-1}\mid z_k,c)
=-\frac{1}{2d}\sum_{j=1}^{d}
\left[
\frac{(z_{k-1,j}-\mu_{\theta,j})^2}{\eta_k^2}
+\log(2\pi\eta_k^2)
\right].
\label{eq:app-log-density}
\end{equation}
The ratio used in the clipped surrogate is
\begin{equation}
\widetilde\rho_{i,k}
=\exp\!\left(
\ell_\theta-\ell_{\theta_{\mathrm{old}}}
\right),
\label{eq:app-ratio}
\end{equation}
with both densities evaluated at the same recorded transition.
This is the $1/d$ power of the corresponding joint density ratio.
Behavior-policy log densities remain fixed during all four
optimization passes on a group.

The resulting loss is
\begin{align}
\mathcal{L}_{\mathrm{PWM}}
={}&-\frac{1}{GK}\sum_{i=1}^{G}\sum_{k=1}^{K}
\min\!\left(
\widetilde\rho_{i,k}A_i,
\operatorname{clip}(\widetilde\rho_{i,k},
1-\varepsilon_{\mathrm{clip}},
1+\varepsilon_{\mathrm{clip}})A_i
\right)
\nonumber\\
&+\frac{\lambda}{GK}\sum_{i=1}^{G}\sum_{k=1}^{K}
\frac{
\|\mu_\theta(s_{i,k})-\mu_{\theta_0}(s_{i,k})\|_2^2
}{d}.
\label{eq:app-loss}
\end{align}
The anchor compares transition means at the same denoising state.
The frozen-base mean is obtained with the adapter disabled.

\subsection{Adapter Checkpoint Averaging}
\label{app:soup}

We uniformly average the corresponding LoRA parameter tensors
from the eight epoch checkpoints of each scene's adaptation run.
For a linear layer with pretrained weight $W_0$, write its adapted
weight as $W_0+(\alpha/r)BA$.
The delivered layer uses
\begin{equation}
\overline A=\frac{1}{8}\sum_{e=1}^{8}A^{(e)},
\qquad
\overline B=\frac{1}{8}\sum_{e=1}^{8}B^{(e)},
\qquad
W_{\mathrm{PWM}}
=W_0+\frac{\alpha}{r}\overline B\,\overline A.
\label{eq:app-adapter-average}
\end{equation}
Thus, averaging operates on the trainable adapter tensors;
the resulting update is the product of their averages.
The final model retains one rank-$8$ adapter.
All eight checkpoints receive equal weight, and checkpoint
construction uses no held-out test frames.

\section{Reward Definitions}
\label{app:rewards}

Let $x_{1:L}$ be the reference support window and $y_{1:L}$
the generated video used for reward measurement, with $L=29$.
We specify the four components of the shared reward below.
All reference quantities are computed from the current support
window.

\subsection{Feature Extraction and Temporal Aggregation}
\label{app:reward-preprocessing}

We use DINOv2-small for visual features, AlexNet-based LPIPS
for perceptual distance, and RAFT-small~\citep{teed2020raft} for optical flow.
Images are resized to $224\times224$ using bilinear interpolation
with \texttt{align\_corners=False}.
DINOv2 inputs use ImageNet normalization with channel means
$(0.485,0.456,0.406)$ and standard deviations
$(0.229,0.224,0.225)$.
LPIPS and RAFT inputs are mapped to $[-1,1]$.

Frame-wise appearance scores are aggregated as
\begin{equation}
\mathcal{A}(a_{1:L})
=\frac{0.7}{L}\sum_{t=1}^{L}a_t
+0.3\min_{1\leq t\leq L}a_t.
\label{eq:app-temporal-aggregation}
\end{equation}
The minimum term gives additional weight to the lowest-scoring
frame within the window.

\subsection{Reward Components}
\label{app:reward-components}

\paragraph*{Scene feature fidelity.}
Let $f_{\mathrm{cls}}(u)$ and $f_p(u)$ denote the DINOv2 CLS
feature and the feature at patch position $p$, respectively.
With $P$ patches, we compute
\begin{align}
q_t={}&0.3\cos\!\left(
f_{\mathrm{cls}}(y_t),f_{\mathrm{cls}}(x_t)
\right)
\nonumber\\
&+\frac{0.7}{P}\sum_{p=1}^{P}
\cos\!\left(f_p(y_t),f_p(x_t)\right),\\
R_{\mathrm{gt}}={}&\mathcal{A}(q_{1:L}).
\label{eq:app-reward-gt}
\end{align}
Patch features are paired at the same grid positions, combining
global visual agreement with spatially localized feature agreement.

\paragraph*{Perceptual appearance.}
The perceptual component is
\begin{equation}
a_t^{\mathrm{lpips}}
=1-\operatorname{LPIPS}(y_t,x_t),
\qquad
R_{\mathrm{lpips}}
=\mathcal{A}(a_{1:L}^{\mathrm{lpips}}).
\label{eq:app-reward-lpips}
\end{equation}
It assigns higher reward to frames with smaller perceptual
distance from the reference.

\paragraph*{Initial-frame consistency.}
For the independent support windows used in adaptation,
the reference visual state is the window's ground-truth
first frame:
\begin{equation}
R_{\mathrm{cross}}
=\cos\!\left(
f_{\mathrm{cls}}(y_1),f_{\mathrm{cls}}(x_1)
\right).
\label{eq:app-reward-cross}
\end{equation}
This component anchors the generated window to its initial
condition.

\paragraph*{Motion magnitude matching.}
Let $F(u_t,u_{t+1})_p$ be RAFT's final flow prediction at pixel
$p$, and let $\Omega$ denote the resized pixel grid.
Define
\begin{align}
M(u_{1:L})
&=\frac{1}{(L-1)|\Omega|}
\sum_{t=1}^{L-1}\sum_{p\in\Omega}
\|F(u_t,u_{t+1})_p\|_2,\\
R_{\mathrm{dyn}}
&=\operatorname{clip}\!\left(
1-\frac{
|M(y_{1:L})-M(x_{1:L})|
}{
M(x_{1:L})+10^{-3}
},
-1,1
\right).
\label{eq:app-reward-dyn}
\end{align}
This term rewards agreement in the average motion magnitude,
penalizing both larger and smaller magnitudes relative to the
support video.
The four components enter the advantage calculation in
Appendix~\ref{app:advantages} with weights
$(0.50,0.30,0.10,0.10)$.

\section{Evaluation Details}
\label{app:evaluation}

Our evaluation measures the benefit of adapting a pretrained world
model to an individual scene.
We compare each adapted model with native Yume on the same held-out
continuation, then summarize both the magnitude and the distribution
of scene-level benefits.
This appendix specifies the paired evaluation, aggregation rules,
and scoring implementation.

\subsection{Paired Scene Continuations}
\label{app:paired-evaluation}

Each scene provides a support trajectory for adaptation and a later,
non-overlapping test trajectory for evaluation.
PWM and Native use the same initial visual context, control sequence,
native generation geometry, decoding configuration, and $145$-frame
evaluation horizon.
Native is evaluated with the scene adapter disabled.
In the SFT comparison, the SFT adapter is evaluated under the same
conditions.

We use the paired random seeds $\mathcal{Q}=\{42,123,777\}$ with
deterministic inference for each seed.
Predictions are scored against the corresponding ground-truth test
frames over the full horizon.
The scene is the unit of aggregation: seed-level scores are averaged
within a scene before results are combined across scenes.

\subsection{Measuring Customization Benefit}
\label{app:benefits}

Using Native as a common reference measures the value added by each
adaptation method to the pretrained model.
Let $S_{s,q}^{(m)}$ denote the customization-fidelity score of method
$m$ on scene $s$ with seed $q$, where lower scores indicate closer
agreement with the reference.
The scene-level score is
\begin{equation}
\overline{S}_s^{(m)}
=\frac{1}{|\mathcal{Q}|}
\sum_{q\in\mathcal{Q}}S_{s,q}^{(m)}.
\label{eq:app-scene-score}
\end{equation}
For $m\in\{\mathrm{PWM},\mathrm{SFT}\}$, define the absolute benefit
and relative benefit as
\begin{align}
B_s^{(m)}
&=\overline{S}_s^{(\mathrm{Native})}
  -\overline{S}_s^{(m)},
\label{eq:app-absolute-benefit}\\
G_s^{(m)}
&=100\,
\frac{\overline{S}_s^{(\mathrm{Native})}
      -\overline{S}_s^{(m)}}
     {\overline{S}_s^{(\mathrm{Native})}}.
\label{eq:app-relative-benefit}
\end{align}
The relative benefit is defined for positive Native scores and is
expressed as a percentage.
Positive values indicate improvement; negative values indicate
degradation.
We first average the scores over seeds and then compute the relative
benefit using that scene's Native score.
This expresses adaptation gains relative to the pretrained model's
error on each target scene.

\subsection{Aggregation Across Scenes}
\label{app:benefit-distribution}

For a domain containing the scene set $\mathcal{D}$, we assign equal
weight to every scene:
\begin{equation}
\overline{G}_{\mathcal{D}}^{(m)}
=\frac{1}{|\mathcal{D}|}
\sum_{s\in\mathcal{D}}G_s^{(m)}.
\label{eq:app-domain-benefit}
\end{equation}
Overall aggregation applies the same rule to the union of the domain
scene sets.
PWM and SFT are summarized over the same paired scene set.

To define conditional summaries of degradation, consider the negative
portion of each method's scene-level benefit distribution. 
Let
\begin{equation}
\mathcal{N}_m
=\{s\in\mathcal{D}:G_s^{(m)}<0\}.
\end{equation}
For nonempty $\mathcal{N}_m$, the conditional mean and median are
\begin{align}
G_{\mathrm{neg,mean}}^{(m)}
&=\frac{1}{|\mathcal{N}_m|}
\sum_{s\in\mathcal{N}_m}G_s^{(m)},\\
G_{\mathrm{neg,median}}^{(m)}
&=\operatorname{median}_{s\in\mathcal{N}_m}G_s^{(m)}.
\label{eq:app-negative-benefits}
\end{align}
These values retain their negative sign; values closer to zero
indicate shallower degradation.
The subset size $|\mathcal{N}_m|$ counts scenes with negative benefit
and is determined separately for each method.
If no scene has negative benefit, these conditional statistics
are undefined.
The mean benefit and these conditional summaries jointly describe
the scale of gains and regressions across scene customization tasks.

\subsection{Scoring Implementation}
\label{app:score-implementation}

\paragraph*{Time-aligned appearance agreement.}
We extract CLS features with \texttt{facebook/dinov2-base}, using
its associated image processor.
Let $\phi(u)$ denote the $\ell_2$-normalized feature of frame $u$.
For generated frames $g_{1:T}$ and ground-truth test frames
$x_{1:T}$, with $T=145$, we compute
\begin{equation}
D_{\mathrm{pair}}
=\frac{1}{T}\sum_{t=1}^{T}
\left(1-\phi(g_t)^\top\phi(x_t)\right).
\label{eq:app-paired-distance}
\end{equation}
Each frame is compared with the reference at the same timestep and
contributes equally.
This evaluates scene appearance as the controlled continuation unfolds.
The evaluation uses DINOv2-base CLS features; the training reward
in Appendix~\ref{app:rewards} uses DINOv2-small CLS and patch features.

\paragraph*{Entity-class consistency.}
The entity component measures the introduction of selected entity
classes relative to the initial test frame.
We use \texttt{openai/clip-vit-base-patch32} with its associated
processor and the fixed prompt pairs in
Table~\ref{tab:app-clip-prompts}.
Let $v(u)$ be the normalized image embedding and $e_c^+,e_c^-$
the normalized positive and negative text embeddings for class $c$.
The two-prompt CLIP score is
\begin{equation}
p_c(u)
=\frac{\exp\!\left(100\,v(u)^\top e_c^+\right)}
{\exp\!\left(100\,v(u)^\top e_c^+\right)
+\exp\!\left(100\,v(u)^\top e_c^-\right)}.
\label{eq:app-entity-score}
\end{equation}

\begin{table}[htbp]
\centering
\small
\caption{Fixed CLIP prompt pairs used for entity-class scoring.}
\label{tab:app-clip-prompts}
\begin{tabular}{
@{}p{0.12\linewidth}p{0.35\linewidth}p{0.42\linewidth}@{}
}
\toprule
Class & Positive prompt & Negative prompt \\
\midrule
Person
& a photo containing a person
& a photo of an empty scene with no people \\
Vehicle
& a photo containing a car or truck
& a photo of a street with no vehicles \\
Animal
& a photo containing an animal
& a photo of a scene with no animals \\
\bottomrule
\end{tabular}
\end{table}

For
$\mathcal{C}=\{\mathrm{person},\mathrm{vehicle},\mathrm{animal}\}$,
the initial frame determines the known-class indicators
\begin{equation}
b_c=\mathbf{1}[p_c(x_1)>0.30].
\end{equation}
A generated frame is flagged if any class absent under this rule
exceeds the generation threshold:
\begin{equation}
h_t=\mathbf{1}\!\left[
\exists c\in\mathcal{C}:
b_c=0\ \land\ p_c(g_t)>0.50
\right].
\end{equation}
The entity term and final score are
\begin{equation}
D_{\mathrm{hall}}=\frac{1}{T}\sum_{t=1}^{T}h_t,
\qquad
S=D_{\mathrm{pair}}+0.5D_{\mathrm{hall}}.
\label{eq:app-final-score}
\end{equation}
Each frame contributes at most one entity flag, even if multiple
classes trigger it.
The same feature extractors, prompts, thresholds, and score weights
are used for all compared methods.

\end{document}